\documentclass[conference]{IEEEtran}
\IEEEoverridecommandlockouts
\usepackage{cite}
\usepackage{amsmath,amssymb,amsfonts}
\usepackage[ruled,vlined]{algorithm2e}
\usepackage{graphicx}
\usepackage{subfigure}
\usepackage{booktabs} 
\usepackage{adjustbox}
\usepackage{textcomp}
\usepackage{xcolor}
\def\BibTeX{{\rm B\kern-.05em{\sc i\kern-.025em b}\kern-.08em
    T\kern-.1667em\lower.7ex\hbox{E}\kern-.125emX}}

\begin{document}

\title{Semi-supervised and Deep learning Frameworks for Video Classification and Key-frame Identification
}\vspace{-0.5cm}
\author{\IEEEauthorblockN{}
\IEEEauthorblockA{Sohini Roychowdhury,\\
\textit{Adjunct Faculty, Computer Engineering } \\
Santa Clara University, email: roych@uw.edu}
}\vspace{-0.5cm}

\maketitle

\begin{abstract}
Automating video-based data and machine learning pipelines poses several challenges including metadata generation for efficient storage and retrieval and isolation of key-frames for scene understanding tasks. In this work, we present two semi-supervised approaches that automate this process of manual frame sifting in video streams by automatically classifying scenes for content and filtering frames for fine-tuning scene understanding tasks. The first rule-based method starts from a pre-trained object detector and it assigns scene type, uncertainty and lighting categories to each frame based on probability distributions of foreground objects. Next, frames with the highest uncertainty and structural dissimilarity are isolated as key-frames. The second method relies on the simCLR model for frame encoding followed by label-spreading from 20\% of frame samples to label the remaining frames for scene and lighting categories. Also, clustering the video frames in the encoded feature space further isolates key-frames at cluster boundaries. The proposed methods achieve 64-93\% accuracy for automated scene categorization for outdoor image videos from public domain datasets of JAAD and KITTI. Also, less than 10\% of all input frames can be filtered as key-frames that can then be sent for annotation and fine tuning of machine vision algorithms. Thus, the proposed framework can be scaled to additional video data streams for automated training of perception-driven systems with minimal training images.
\end{abstract}


\section{Introduction}
\begin{figure}[ht!]\label{system}
\includegraphics[width=3.4in,height=1.2in]{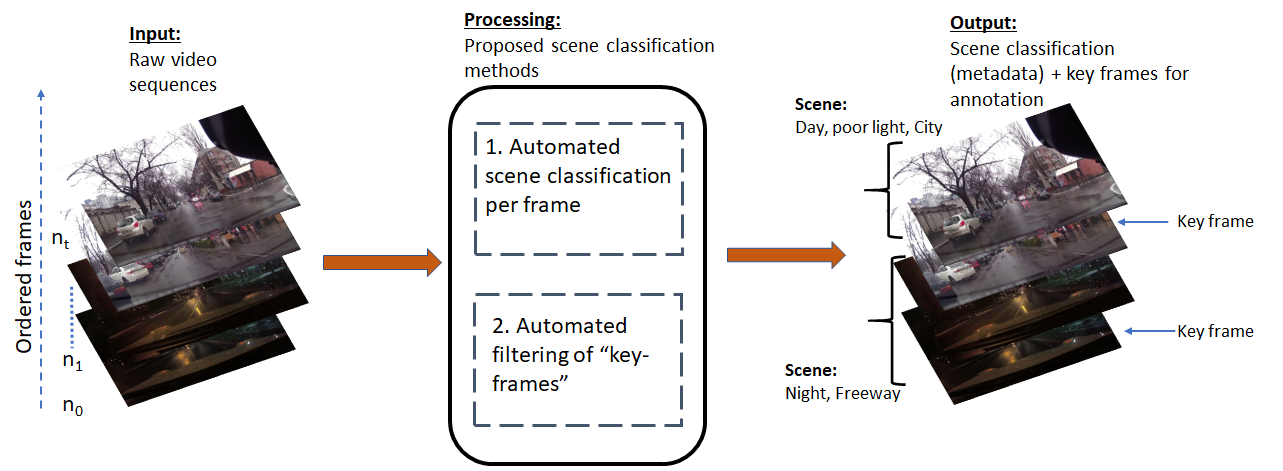} 
\caption{Proposed framework for automated scene classification per frame and for key-frame filtering. Raw video sequences are supplied to either one of the two proposed semi-supervised algorithms such that each video frame gets tagged for lighting conditions (day, night, poor lighting, shadows, low standing sun etc.) and content (city, freeway, pedestrians, parked-cars). Also frame IDs corresponding to highly \textit{uncertain} frames are filtered as key-frames. }\vspace{-0.5cm}
\end{figure}

Computer Vision and scene understanding applications rely on processing large volumes of video data to learn the patterns related to the objects/regions of interest (ROIs) \cite{keyframe}. Therefore, automated video sequence processing systems for efficient frame tagging, storage and retrieval become key components for building such automated machine vision systems. Also, there is a constant need to isolate \textit{key-frames} with high structural variance and good quality from video sequences that can then be used to train machine learning models \cite{videochallenges}. In this work we present two methods that rely on minimal annotated data to automatically classify and tag individual frames from video sequences based on the scene type and lighting conditions. Also, both methods can filter key-frames with varying processing speed and reliability across datasets to identify minimal training datasets for subsequent training of machine vision algorithms. The proposed end-to-end system is shown in Fig. \ref{system}.

Till date, several image-specific metrics have been applied for key-frame selection based on image quality such as: peak signal to noise ratio (PSNR), contrast to noise ratio (CNR, that is sensitive to compression formats), SIFT features \cite{SSIM} and mean squared error. However, these image metric-based methods can only eliminate poor quality image frames with blurriness and poor light. Besides, deep learning-based feature extraction methods in \cite{keyframe} \cite{summary} have been applied to narrow field-of-view image sequences for encoding and video content summarizing tasks. These deep learning methods have shown to encode actions from video sequences only after learning from large volumes of annotated data, which is a costly process altogether. Since annotation costs for video-based applications are highly dependent on training data volumes, there is a need to streamline the process of key-frame identification, and to curate sequences with appropriate metadata. Additionally, the recent work in \cite{keyframedeep} used self-supervised deep-learning representations to isolate key frames from targeted human action-specific tasks only. Motivated by these existing works, we implement a novel two-stage end-to-end process to first automatically classify each frame from video sequences for frame content, uncertainty and lighting conditions for multiple targets across wide-fields of view videos. Next, key-frames are selected as ones with the highest degree of structural uncertainty, dissimilarity and high image quality. 

This paper makes two major contributions. First, we present two methods that require less than 20\% of available video frames to be annotated for automated classification and tagging of outdoor scenes and lighting conditions per frame. The rule-based method has minimal parameters and is more robust to training data variations when compared to the novel deep-learning based method. However, the rule-based method has ten-times more computation time complexity and captures less variety in structural content such as snow accumulations and weather-related variations for key-frames when compared to the deep learning-based method. Second, we present novel key-frame filtering methods that isolate less than 10\% of all frames from video sequences that represent structural variety and high quality for improved scene understanding tasks.

\section{Data and Methods}
The details of the two proposed methods for per-frame scene classification and key-frame filtering are presented below. The first rule-based baseline method utilizes sequential image content information to identify the uncertain key-frames. The second novel deep-learning representation based method requires a critical mass of annotated frames to automatically perform scene classification based on clustering trends per frame. We assess both methods for speed, accuracy of per-frame classification and generalizability across outdoor datasets.  

\subsection{Data}
We implement scene classification and key-frame filtering by combining individual video sequences from public datasets of JAAD \cite{JAAD} and KITTI \cite{KITTItracking}. The manual annotations per sequence will be publicly released for benchmarking purposes. The goal is to identify scene changes and to isolate frames with previously unseen content that can further enhance scene understanding tasks. For example, object detection tasks in night-time sequences are specifically difficult and standard object detection algorithms trained on day-time images do not scale well in such situations. Identifying key-frames where objects such as pedestrians, bicyclists etc. first appear into a scene, or frames where key objects are missed due to poor lighting is crucial to further fine-tuning such object detectors. Other scene understanding tasks that can be further fine tuned by key-frames can be object trajectory prediction, pedestrian intention prediction and landmark detection tasks as shown in \cite{videochallenges}. Each video sequence is manually annotated for the following 3 parameters: 
\begin{itemize}
    \item Time of Day: \{night(1), day(0)\}
    \item Lighting conditions: \{poor(1), good (0)\}
    \item Scene Class: \{city (0), pedestrians (1), freeway (2), parked-cars (3)\}
\end{itemize}

\begin{figure*}[ht]
    \centering
	\subfigure[Image with \{night, good light, city\} from sequence 0024, JAAD]
	{\includegraphics[width=0.45\textwidth, height=0.9in]{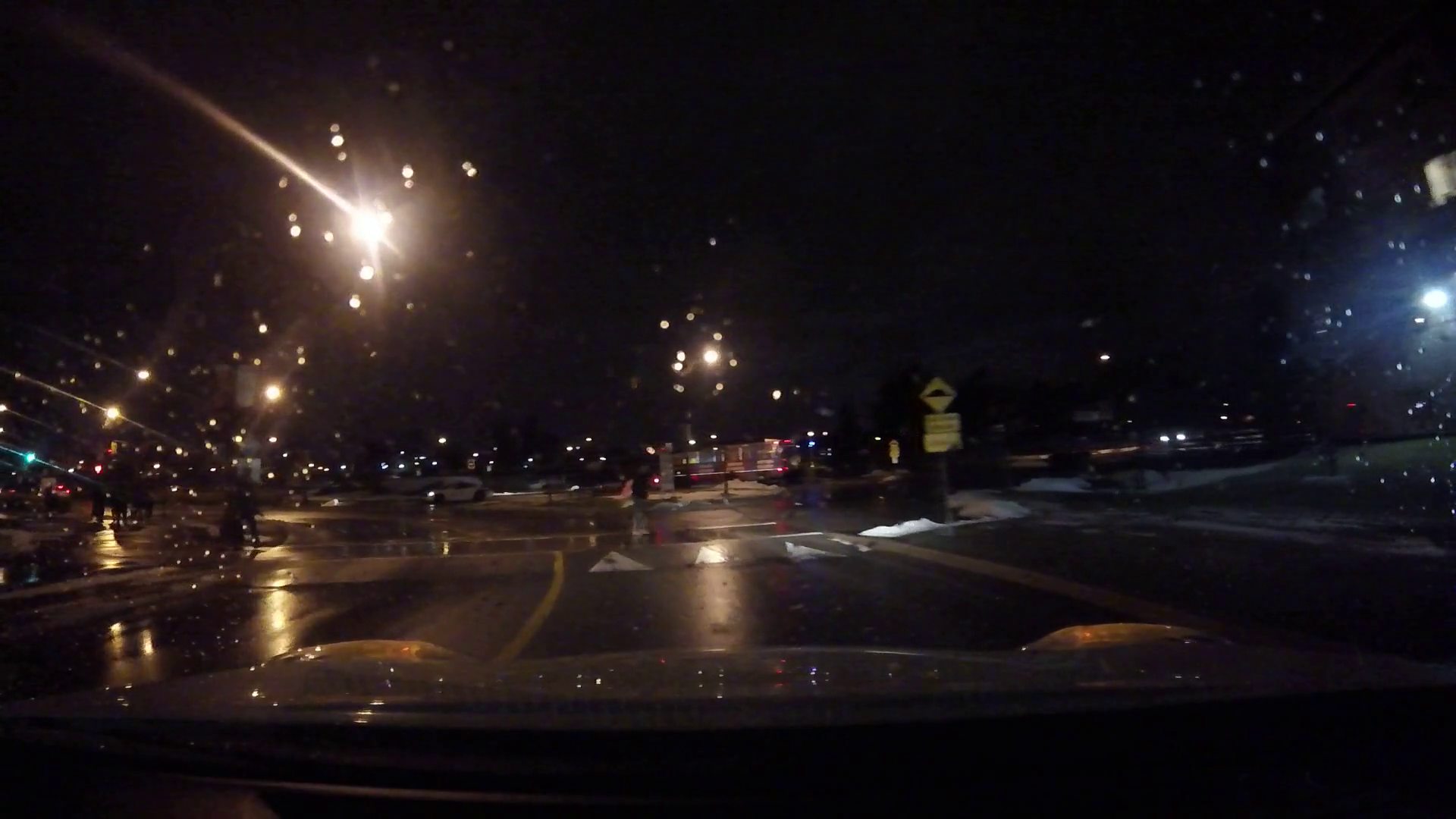}}
	\subfigure[Image with \{day, good light, city\} from sequence 0283, JAAD]
	{\includegraphics[width=0.45\textwidth, height=0.9in]{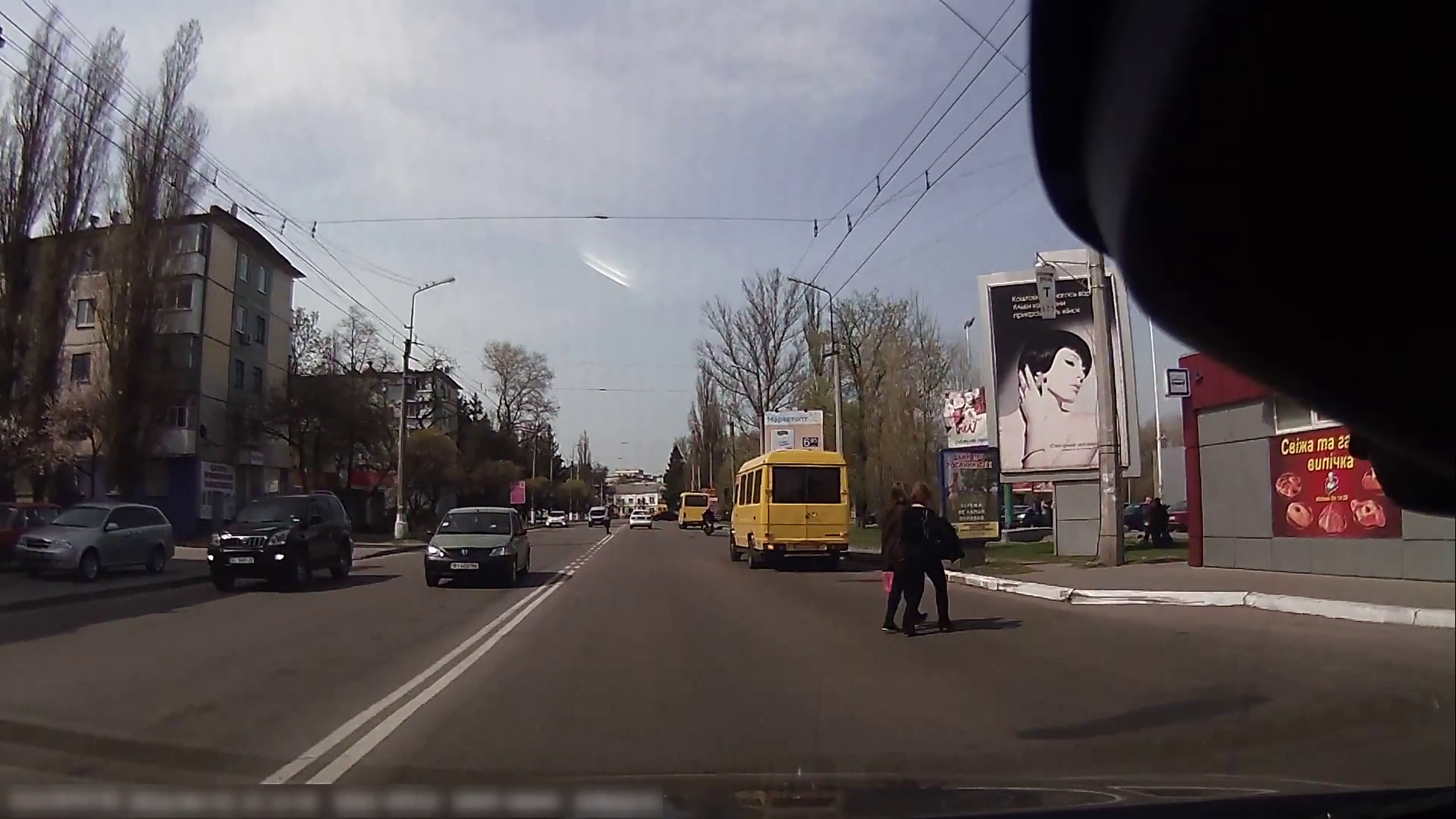}}
	\subfigure[Image with \{day, poor light, city\} from sequence 0280, JAAD]
	{\includegraphics[width=0.45\textwidth, height=0.9in]{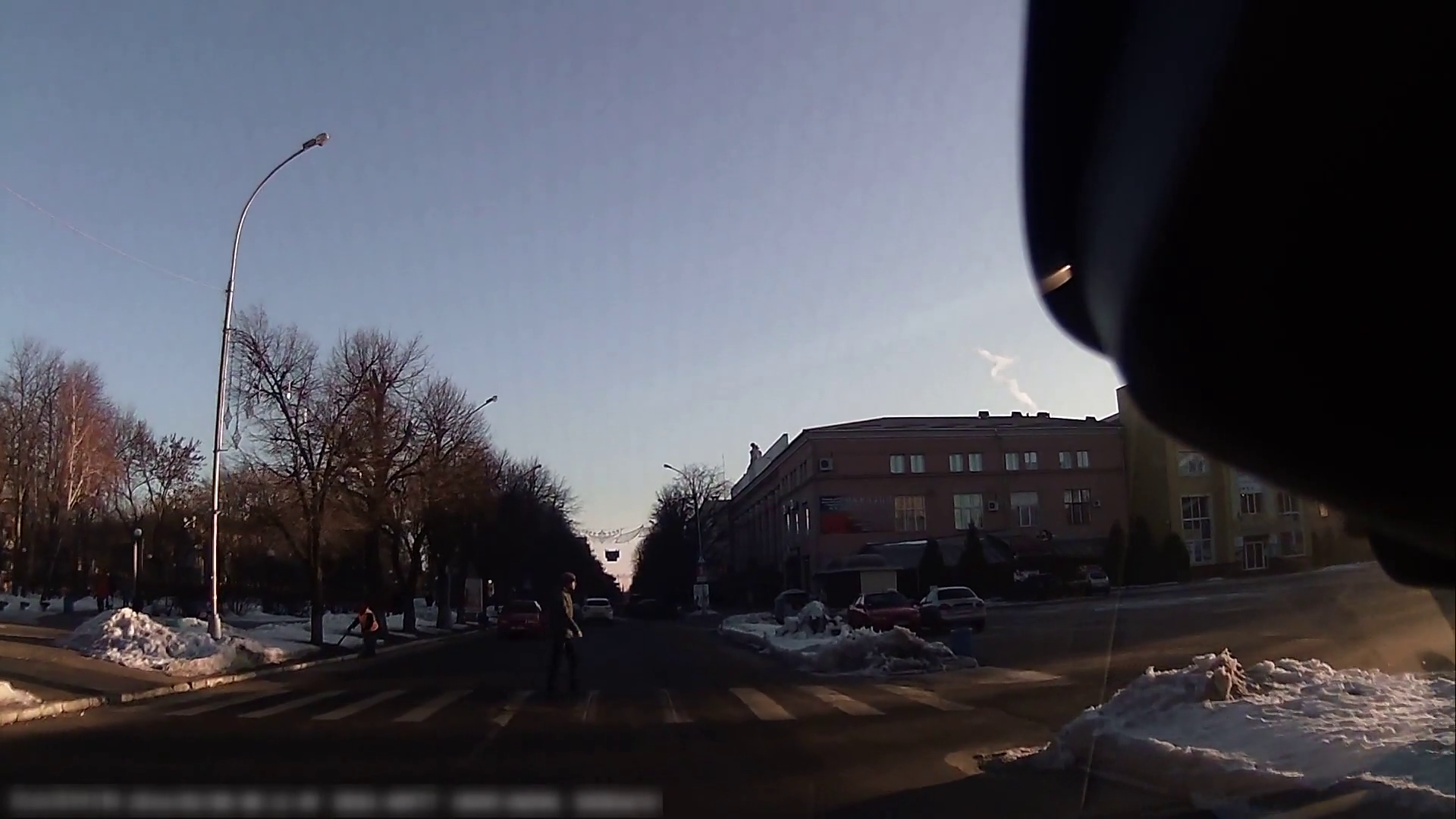}}
	\subfigure[Image with \{day, poor light, city\} from sequence 0279, JAAD]
	{\includegraphics[width=0.45\textwidth, height=0.9in]{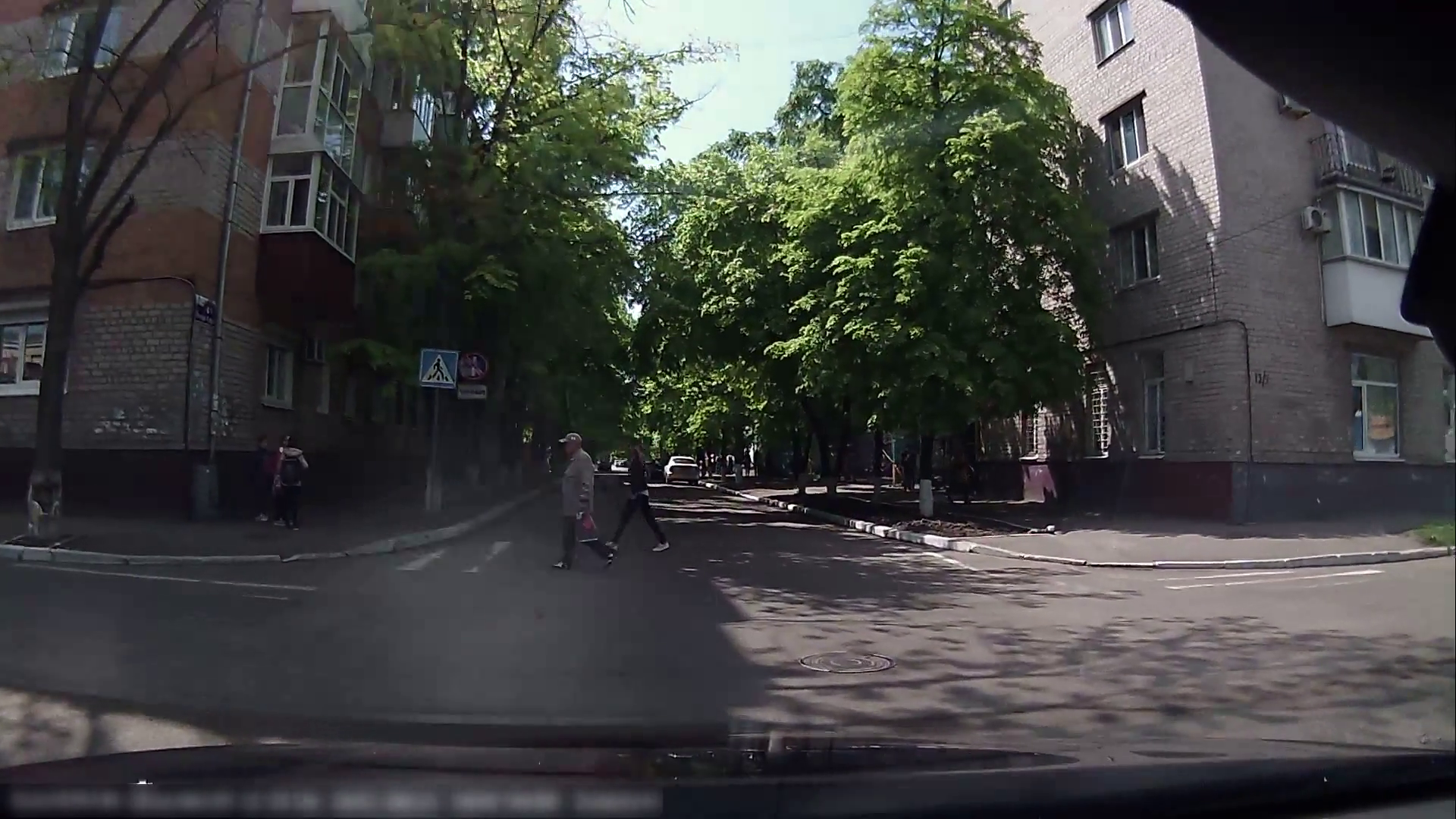}}
	\subfigure[Image with \{day, poor light, pedestrians\}, sequence 0292, JAAD]
	{\includegraphics[width=0.45\textwidth, height=0.9in]{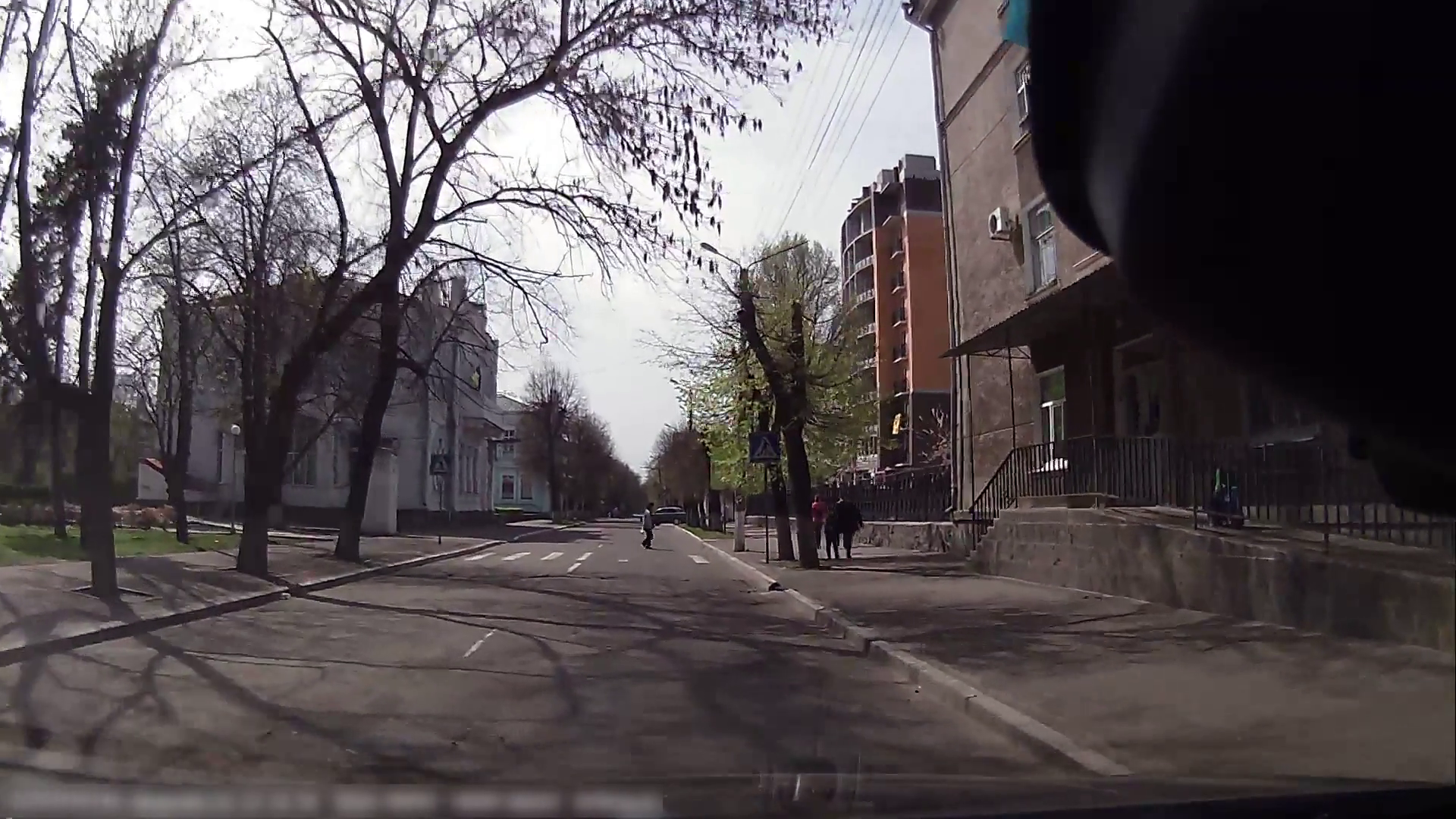}}
	\subfigure[Image with \{day, good light, pedestrians\}, sequence 0295, JAAD]
	{\includegraphics[width=0.45\textwidth, height=0.9in]{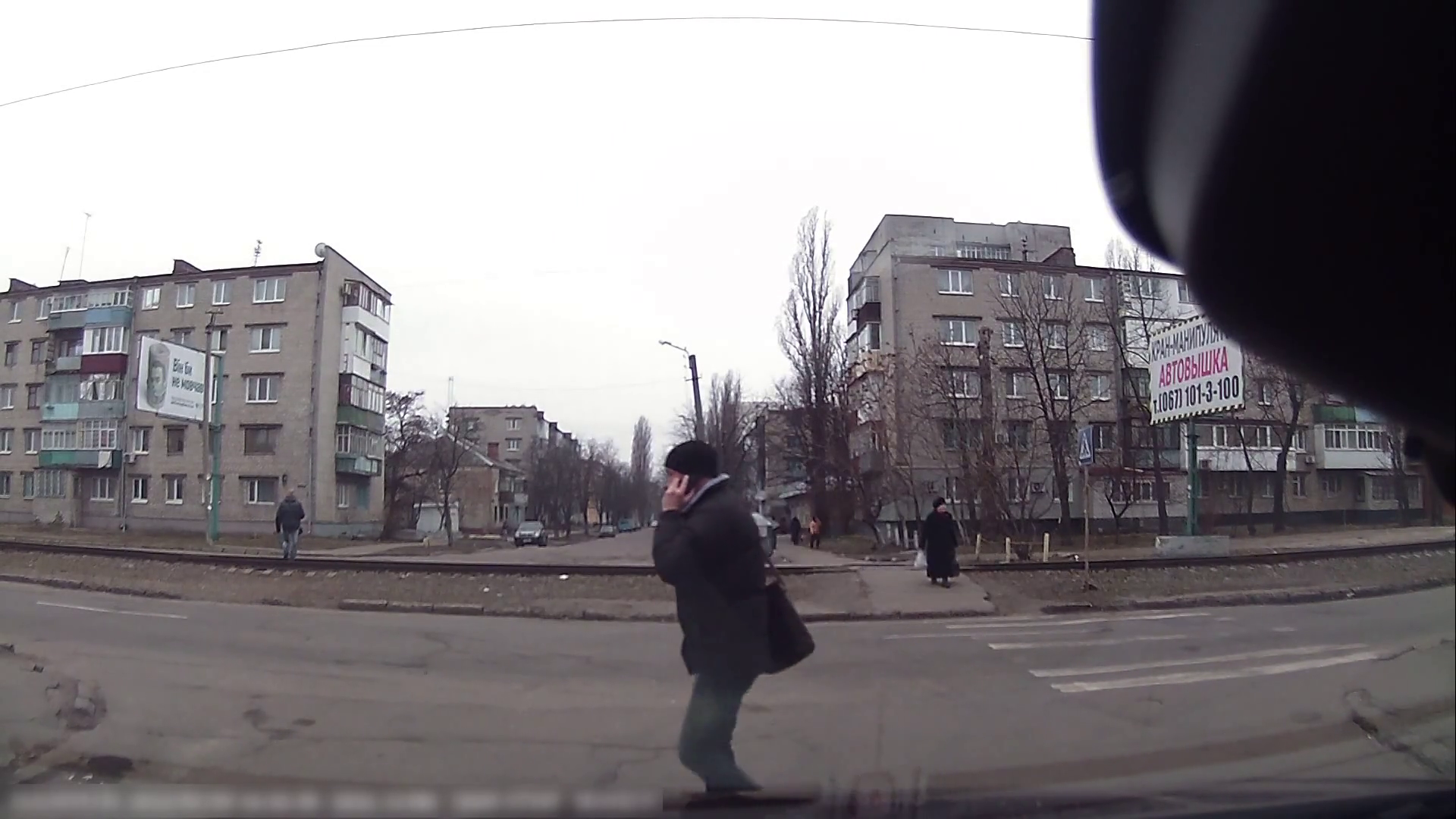}}
	\subfigure[Image with \{day, poor light, freeway\}, sequence 0005, KITTI]
	{\includegraphics[width=0.45\textwidth, height=0.9in]{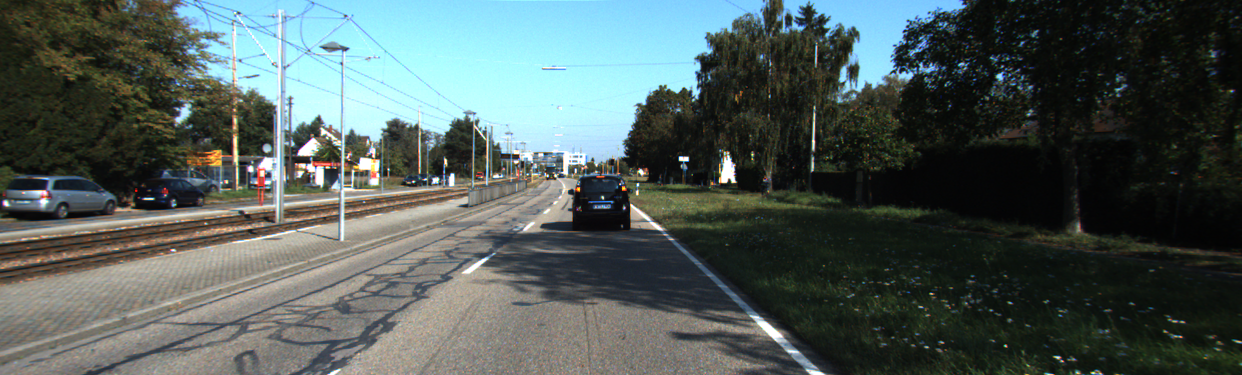}}
	\subfigure[Image with \{day, good light, freeway\} from sequence 0008, KITTI]
	{\includegraphics[width=0.45\textwidth, height=0.9in]{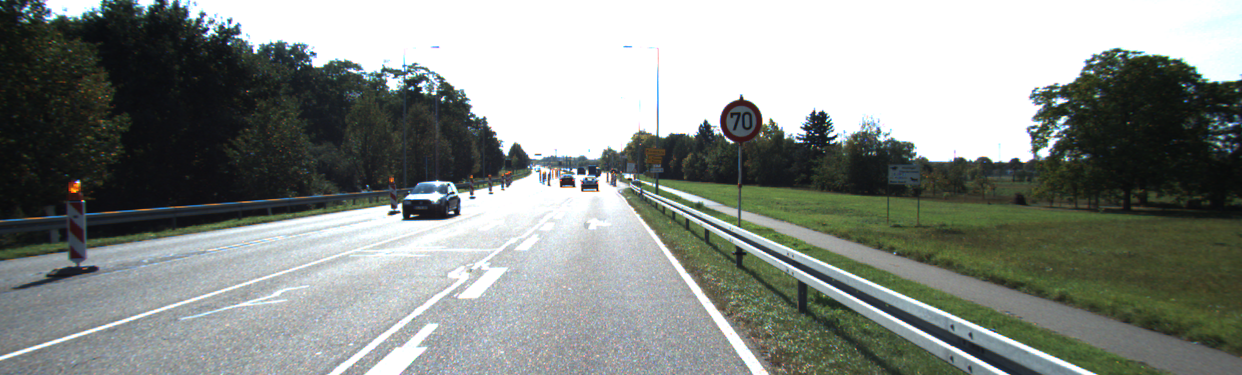}}
	\caption{Examples of Scene categories from the JAAD and KITTI datasets.}\label{examples}
       \vspace{-0.3cm}
\end{figure*}

The first dataset is the sub-sampled sequences from the JAAD data set \cite{JAAD} with 54 video sequences of outdoor scenes with sequence names [0022-0024, 00273-0324]. These video sequences are selected since they represent varying lighting and weather conditions, such that each video sequence contains an average of 237 frames per sequence. The JAAD dataset is acquired at 30 frames-per-second and each image has dimensions [1080 x 1920] pixels per frame. Combining all video frames together results in 6101 total frames with annotated pedestrians only. Each JAAD scene is manually annotated to belong to one of the five following categories: [\{day, good light, city\}, \{day, good light, pedestrians\}, \{day, poor light, pedestrians\}, \{day, poor light, city\}, \{night, good light, city\}] with representations [10, 7.6, 2.16 , 77.93, 4.92, 7.38]\% for each category/class, respectively. We observe a major scene category corresponding to daytime city scenes with shadows and poor lighting. To isolate a minimal training dataset with similar distributions as test data, we select all frames from the video sequences [0022, 0023, 0273, 0286, 0290, 0292, 0295, 0310, 0303, 0307, 0312] as training data and all remaining frames become test data. Examples of scene categories are shown in Fig. \ref{examples}.

The second dataset is the KITTI tracking dataset for 2D objects \cite{KITTItracking} that has a variety of video-sequences of outdoor scenes of varying lengths. These video sequences are acquired at a frame-rate of 10 frames-per-second and the images are cropped frames with [1382 x 512] pixels per frame. Here, we use the 21 video sequences with names [0000-0020] from the training folder with 8008 total frames. Each video sequence belongs to one or more of the following five categories: [\{day, good light, freeway\}, \{day, poor light, parked-cars\}, \{day, poor light, city\}, \{day, poor light, freeway\}, \{day, poor light, pedestrians\}] with representations [8.24, 35.35, 32, 20.2, 4.42]\% per category, respectively. Here, the category \textit{parked-cars} refers to videos with several parked cars, either in parking lots or street parking. The minimal training dataset with similar distribution of scenes as test data, is selected as all frames from the sequences [0000, 0001, 0005, 0006, 0017]. All frames from the remaining sequences are combined to create the test video sequence.

\subsection{Frame Selection and Quality Detection Modules}\label{sel}
Scene understanding tasks typically require learning from a variety of training samples. Thus, selection of frames that are significantly dissimilar in structural content and high in quality is necessary. One key building block for our scene categorization models is a frame structural selector and quality selector module that eliminates consecutive frames from a sequence that are structurally similar to others and that may be blurry.

In the algorithm \ref{algo}, we implement such a frame selection module which starts iterating frames ($j$) from the first frame ($I_0$) as a reference frame and compares subsequent frames to the reference frame based on the structural similarity index metric (SSIM, $s$) \cite{SSIM}. As soon as a frame with $SSIM<\alpha$, where $\alpha \in [0,1]$ is encountered, this new frame becomes the new reference frame ($I_{ref}$), gets added to selected frame set ($F_S$) and the process continues till the last frame ($I_l$). Since SSIM within a frame neighborhood are compared, the computational complexity remains relatively linear.
\begin{algorithm}[ht!]
\SetAlgoLined
\KwOut{Selected Frames $F_S=\{I_k\}$}
 \KwIn {Video frames:$\{I_l\}$, $F_S=\{\phi\}$}
 Initialization\;
 $I_{ref} \leftarrow I_0$, step$\leftarrow$ 2, s$\leftarrow$1, j$\leftarrow$0, $F_S \leftarrow \{I_0\}$\\
 \While{$j<l$}{
  \While{s$> \alpha$}
 {
 j$\leftarrow$ j+step\\
 s$\leftarrow ssim(I_{ref},I_{j})$\\
  }
 $I_{ref} \leftarrow I_j$ \\
 $F_S \leftarrow F_S \cup I_{ref}$ \\
 }
\caption{Frame Structural Selection}\label{algo}
 \vspace{-0.1cm}
\end{algorithm}

The next task is to further filter the structurally dissimilar frames based on image quality to ensure that blurry images are discarded. For this purpose we implement the OpenCV Laplacian of variance function based on existing works in \cite{blur2}. Here, the 2D Laplacian kernel highlights foreground edges, thereby yielding a higher score for pixel variances in well-focused and sharp images. We apply the Laplacian function to grayscale images converted from the video sequences and retain only high quality images with high coefficients. This structural and quality-based frame selection module is used for isolating samples for training the self-supervised deep learning model and for further filtering key-frames from the rule-based method. This module is capable of reducing the number of retained frames from 40-90\% of the initial volume by varying the threshold $\alpha$ between [0.4-0.8] and normalized Laplacian of variance in range [0.4-0.7], respectively.

\subsection{Rule-based Frame Classification: Baseline}
\begin{figure*}[ht!]
    \centering
    \includegraphics[width=5.8in, height=1.7in]{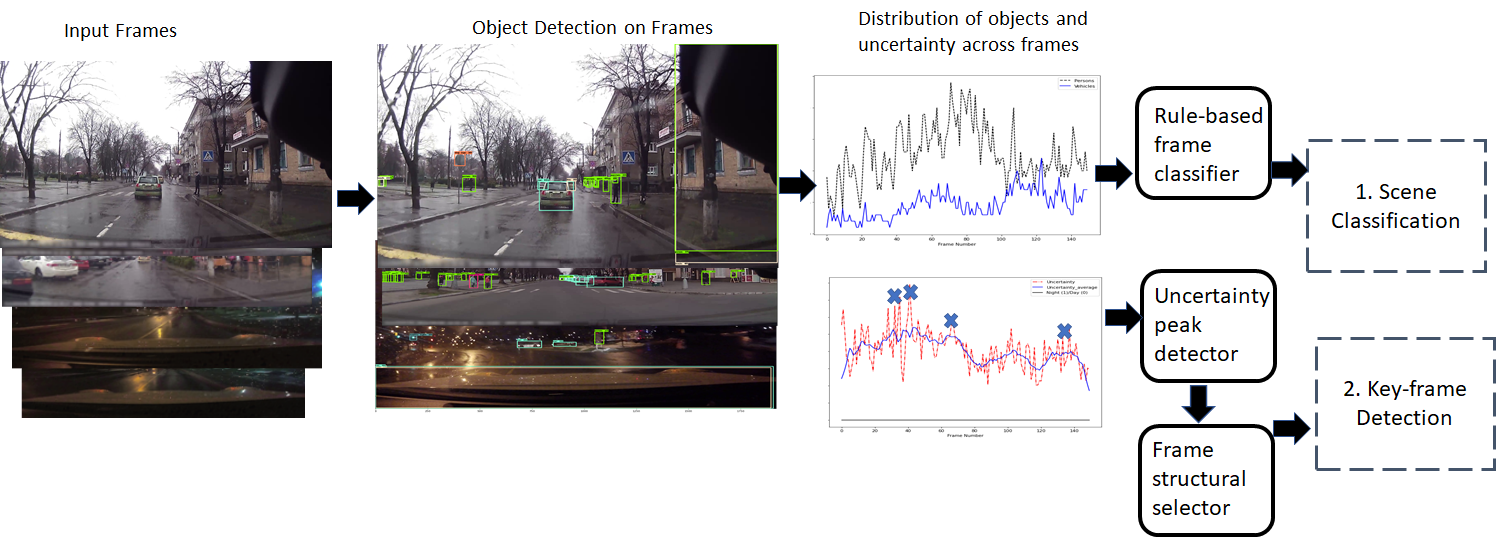}
    \caption{The proposed rule-based model to classify video frames starts with a pre-trained general purpose object detector. Next, the number of foreground objects and probability distributions of objects is used to classify each frame. The most uncertain frames are selected using a peak detector as key-frames.}
    \label{base}
    \vspace{-0.5cm}
\end{figure*}

As a baseline frame classification model with minimal training data requirements, we implement a rule-based method that relies on pre-trained structural detectors to estimate reliable content and relative unreliability or noise per frame. This baseline model that requires minimal training data is shown in Fig. \ref{base}. First, a pre-trained general purpose object detector or semantic segmentation model is used to extract the reliable foreground objects of interest per image. For outdoor images, this can be the number of pedestrians, bicyclists, vehicles, trains etc. Next, based on the number of foreground objects and the distribution in the probability of objectness per frame, each image can be classified for lighting and scene-related metadata. The unreliability/uncertainty of object bounding boxes per frame can then be utilized to screen for key-frames. 

\begin{figure}[btp]
\includegraphics[width=3.4in,height=1.5in]{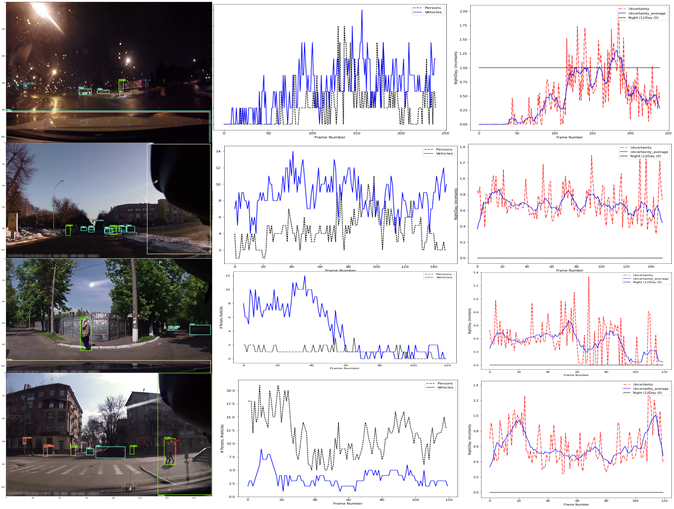} 
\caption{Examples of images with persons (P, blue), vehicles (V, red), uncertainty (U, red dots) and background light (B, black) for frames from JAAD datasets sequences 0023, 0280, 0289, and 0290, respectively. City images show non zero combination of $\{P, V\}$, and images shown here with poor lighting conditions have high uncertainty ($U>0.5$).}\label{metrics}
\end{figure}

Here, we implement a Faster-RCNN model \cite{faster} pre-trained on the MS-COCO dataset to extract foreground object bounding boxes and their objectness probabilities ($r$) for all objects with probabilities greater than 0.2. We retain only the relevant foreground objects in terms of the number of pedestrians and bicyclists ($P$), vehicles such as cars, trucks, buses, airplanes and motorcycles ($V$) and background lighting i.e. night vs. day ($B$). Next, the uncertainty in each frame is estimated in terms of the coefficient of variation ($U= \frac{\sigma_r}{r_{med}}$), where, $\sigma_r$ represents the standard deviation in all relevant foreground object detection probabilities and $r_{med}$ represents the median of the foreground object probabilities. A higher $U$ would imply high variations in foreground object probabilities and therefore captures uncertainties in frames corresponding to poor lighting, shadows, low standing sun etc. Other known methods to estimate uncertainty per frame can be estimating intersection over union for multiple detections with dropout applied at test time. Examples of variations in content ($P,V, B, U$) for varying scenes are shown in Fig. \ref{metrics}.

Once the content of each frame is estimated, the impact of neighboring frames in sequence is computed by averaging the uncertainty for 10 frames within a sliding window. So, for each frame, the mean of 5 previous and 5 subsequent frame uncertainties is computed as the mean uncertainty ($\bar{U}$). Finally, the following frame classifications are made based on the content.
\begin{itemize}
    \item \{$P>V, V<3, B=1$\}, class $\rightarrow$ \{pedestrians, good light, night\}
    \item \{$P>V, V<3, B=0, \bar{U}<=\beta$\}, class $\rightarrow$ \{pedestrians, good light, day\}
    \item \{$P>V, V<3, B=0, \bar{U}>\beta$\}, class $\rightarrow$ \{pedestrians, poor light, day\}
     \item \{$P=0, V>0, B=1$\}, class $\rightarrow$ \{freeway, good light, night\}
     \item \{$P=0, V>0, B=0, U>\delta, \bar{U}<=\beta$\}, class $\rightarrow$ \{freeway, good light, day\}
    \item \{$P=0, V>0, B=0, U>\delta, \bar{U}>\beta$\}, class $\rightarrow$ \{freeway, poor light, day\}
     \item \{$P=0, V>0, B=0, U<\delta, \bar{U}<\beta$\}, class $\rightarrow$ \{parked cars, poor light, day\}
     \item \{$P>=0, V>=0, B=1$\}, class $\rightarrow$ \{city, good light, night\}
    \item \{$P>=0, V>=0, B=0, \bar{U}<=\beta$\}, class $\rightarrow$ \{city, good light, day\}
    \item \{$P>0, V>=0, B=0, \bar{U}>\beta$\}, class $\rightarrow$ \{city, poor light, day\}
 \end{itemize}
For any other combination of $P,V,B,\bar{U}$, the \textit{Other} category is assigned, signifying an unknown combination that can be due to indoor parking areas, tunnels etc. We implement the [1024x1024] pixel Faster-RCNN model from Tensorflow hub and observe that the test time per frame  ranges between [0.15-0.17] seconds. The parameters $\beta$ and $\delta$ are estimated using 5-fold cross validation to identify the parameter in range [0,1] that maximizes weighted accuracy of per-frame classification.

Finally, key frames are identified as frames that cause peaks in relative objectness uncertainties. Here, a peak detector is applied to the $U$ metric across frames, to isolate frames that have sudden variations in object probabilities with respect to its neighborhood frames and that have uncertainty greater than $\eta$. Here, $\eta$ can be varied  in range [0.5-1] to control for the number of filtered key frames with high uncertainty. Once the key-frames are identified, they are further subjected to the frame and quality selector module to ensure minimal structural similarity and high quality for the finally filtered key-frames.

\subsection{Label Propagation with simCLR Representations: Semi-supervised}\label{filter}
\begin{figure*}[ht!]
    \centering
    \includegraphics[width=5.4in, height=1.8in]{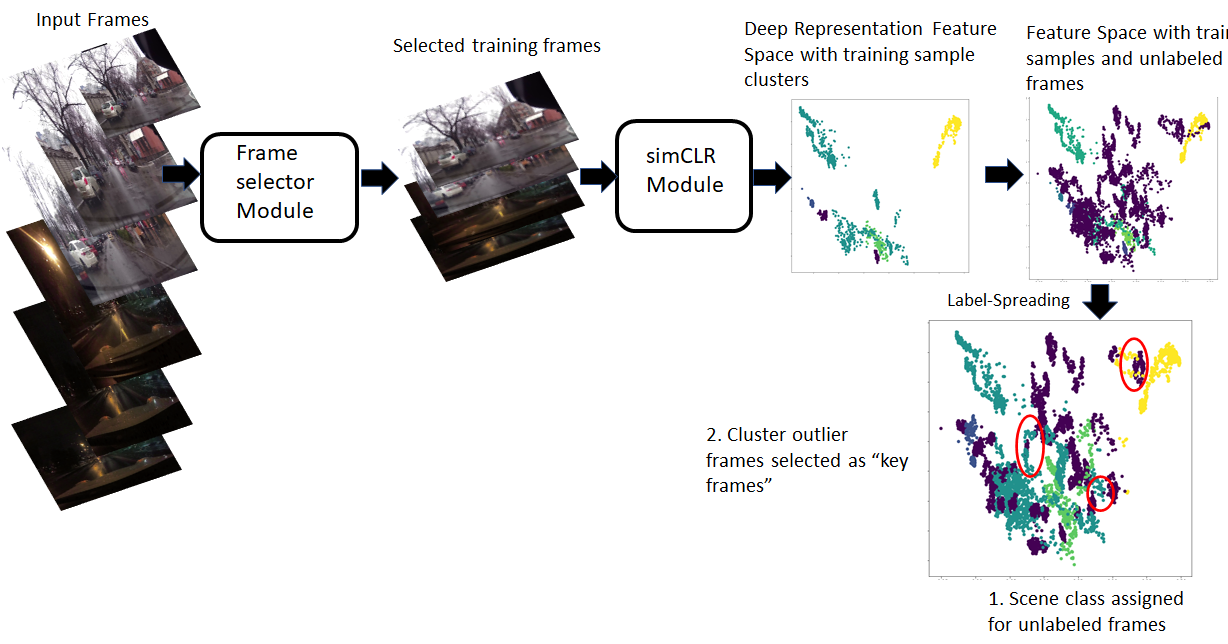}
    \caption{The proposed simCLR-based scene classification and key-frame filtering method starts with a subset of selected training frames for fine-tuning a self-supervised deep learning encoder model. Once the best representation feature space is identified, a semi-supervised approach of label spreading is used to label all test frames. Frames that lie on decision boundaries shown in the red circles are filtered as key-frames.}\label{simCLR}
    \vspace{-0.5cm}
\end{figure*}

As an active learning framework, we propose a semi-supervised method to cluster scene frames based on their deep learning representations. The goal is to assign labels to previously unlabelled frames and to identify samples/frames that lie around cluster boundaries, which are difficult to assign cluster labels, as key-frames. The proposed framework is shown in Fig. \ref{simCLR}.

We begin by implementing a self-supervised simCLR \cite{simclr} model with a ResNet backbone that identifies cropped and modified versions of an image from other images. The simCLR model uses augmented versions of each frame for learning weights as opposed to frame-labels, thereby reducing its dependence on frame annotation. The images needed to train the simCLR model should ideally represent significant structural variations. For this reason, we apply the \textit{frame structural selector} module from Section \ref{sel} on the combined training video sequences from each dataset and extract a training subset to train the simCLR model. At the end of training, all frames from the training sequences get encoded into their deep learning representations. The representations from the last 3 layers prior to classification, corresponding to 128, 256 and 2048 neurons, respectively, are analyzed for the best high-dimensional representation as follows.

All frames from the training video sequences are encoded in terms of the 128, 256 and 2048 neurons, followed by high-dimensional embedding using PCA, t-SNE and UMAP. Next, the labels from the training subset are propagated (by label spreading) through a best fit rbf-kernel to label the remaining training sequence frames. Empirically, we observe that the third from last deep learning layer with 2048 neurons subjected to PCA with 10 dimensions results in the best high-dimensional encoding with highest classification weighted per-frame accuracy on validation data. This process enables encoding each video frame $I$ to an encoded sample $\mathcal{F}(I)$. Next, using encoded training frames ($\mathcal{F}(I_j)$) with the annotated scene class ($Y_j$), we incorporate label-spreading to assign a scene class to each test frame ($\mathcal{F}(I_k)$) using equations \eqref{ls}-(3) based on \cite{labelspread}.

\begin{align}\label{ls}
    \bf{w_{j,k}}=e^{-\gamma||\mathcal{F}(I_j)-\mathcal{F}(I_k)||^2},\\
    S=D^{-\frac{1}{2}}{\bf W}D^{-\frac{1}{2}},\\
    \text{Given, }Y(0)=[y_j, [-1]^{[k]}],\\ \nonumber
    Y(t+1)=\nu SY(t)+(1-\nu)Y(0).
\end{align}
In (1), an rbf-kernel-based distance matrix $\bf{W}$ is computed for each combination of train ($j$) and test images ($k$).  Next, a symmetric normalized Laplacian matrix $S$ is estimated using diagonal matrix $D$, where the $i$th diagonal entry is the sum of the $i$th row of $\bf{W}$ in (2). The initial label vector for the training and test data combined is represented by $Y(0)$ with $j+k$ entries that contains the training labels (with $j$ entries) padded with -1 placeholder labels for the $k$ test samples. Next, the iterative label spreading method continues using (3) till convergence is reached or maximum 30 time steps (t) have been completed. In each new iteration, the label matrix $Y(t+1)$ is updated based on the neighboring training samples. A higher value of $\nu$ in the range [0,1] indicates a higher update in each iteration versus the initial label set $Y(0)$. It is noteworthy that label-spreading using the rbf-kernel has high time complexity for large numbers of training and test samples. However, this process can be applied efficiently for small data batches (up to 10,000 images per train/test batch size). Also, we observe that the simCLR-based frame labeling approach, though more training data dependent when compared to the rule-based method, is significantly faster since classifying all test sequences takes upto 4 minutes on a Windows system with 8GB RAM, 8GB Nvidia 2070 GPU. Additionally, the k-nearest neighbor method can be substituted in (1-2) to reduce the computational time complexity at a cost of up to 5\% reduction in classification accuracies for the data sets here.

Finally, to extract key-frames from test sequences, we implement rbf-label-spreading with varying kernel variances $\gamma$ and learning parameters $\nu$ for 20 runs. We detect the frames/samples that are assigned more than one unique label across the 20 runs. These samples can be considered to lie around cluster boundaries since they get labeled differently across the runs. These cluster boundary samples can then be filtered as key-frames and the volume of filtered frames can be varied by controlling the $\gamma$ and $\nu$ parameters.

\section{Experiments and Results}
We evaluate the performances of the rule-based baseline model and the novel simCLR-based model for test scene classification per frame, for dependence on training data and for key-frame filtering through three experiments. First, we train both models using training frames from JAAD and KITTI data set and test the classification performances per test frame for video sequences from the same dataset. Second, we vary the training datasets for both methods and analyze robustness in the per-frame scene classification performance. Third, we analyze the number of key-frames filtered from each dataset by both methods to assess their agreeability and to qualitatively explain the contents of the filtered frames. The scene classification performances are evaluated in terms of averaged macro and weighted precision, recall, f-score and accuracy across all the classes.

\subsection{Per-frame Scene Classification}
Scene classification performances per frame for the JAAD and KITTI data sets are given in Table \ref{res1}. 
\begin{table*}[ht!]
\caption{Performance of scene classification using the proposed methods.}
\begin{adjustbox}{width=6.4in,center}
\begin{tabular}{|c|c|c c|c|c c|c|}
\hline
Method&Accuracy&Macro Precision&Macro Recall&Macro f-score&Weighted Precision&Weighted Recall&Weighted f-score\\ \hline
Dataset&&Train: JAAD& Test: JAAD&&&&\\\hline
Rule-based&88.5&{\bf 43}&{\bf 46.8}&{\bf 44.2}&{\bf89.3}&88.5&88.8\\ 
simCLR-based&{\bf 93.25}&37.67&39.88&38.71&{87.5}&{\bf 93.25}&{\bf 90.3}\\ \hline
Dataset&&Train: KITTI& Test: KITTI&&&&\\ \hline
Rule-based&{\bf72.2}&{\bf52.6}&47.3&46.6&{\bf69.5}&{\bf72.2}&{\bf69.1}\\
simCLR-based&66.4&45.0&{\bf54.8}&{\bf48.8}&52.3&66.4&57.7\\ \hline
\end{tabular}
\label{res1}
\end{adjustbox}
\end{table*}
Here, we observe that for the JAAD dataset, if all the scene categories were equally likely, the macro classification performances of the rule-based method would be better than the simCLR-based method. However, considering consistent scene category distributions from the train to test scenes, the simCLR-based method has superior weighted accuracy for test frame classification. It is noteworthy that for the JAAD dataset, there exists a class imbalance with a large number of frames belonging to \{day, poor light, city\} category. Thus, over prediction for this class results in higher weighted classification performance. 

On the other hand, for the KITTI dataset, the scene category distribution is relatively uniform. The video sequences in this dataset are longer than JAAD and some sequences contain parked cars in residential areas followed by merging to a freeway in the same sequence. Also, there are sequences that contain parked cars, and city scenes together. Therefore, manual scene classification per sequence may not align with automated per-frame classifications. For the KITTI dataset, in Table \ref{res1}, we observe that the rule-based method has better weighted classification while the simCLR method has better macro classification performances. This occurs because the rule-based method relies on a task specific object detector for foreground and uncertainty detection and is therefore able to classify city scenes from parked cars and pedestrians more efficiently. The simCLR-based method however results in significant mis-classifications between parked-cars, city and freeway scenes since there are significant structural similarities between the frames of all these classes. Also, from Table \ref{res1} we observe that for the KITTI dataset, the longest test video sequence 0020 corresponding to a freeway with 837 frames is misclassified as city due to the foliage along the freeway. This misclassification significantly reduces the overall classification accuracy for the KITTI dataset when compared to JAAD.

\subsection{Robustness Analysis to Training Data}
In this experiment we analyze the generalizability of the two scene classification methods to new use-cases. Here, we vary the training dataset and analyze the scene classification performances per frame. It is noteworthy that the rule-based method is intuitively more generalizable for outdoor scene classification since it relies on a pre-trained foreground object detector and requires minimal parameterization for $\gamma$ and $\delta$ only. Next, we train the semi-supervised simCLR-based label-spreading model with labeled frames from the other dataset. For this analysis, the combined training and test datasets from JAAD and KITTI comprise 8 unique scene categories and only 2 overlapping categories corresponding to \{day, poor light, city\} and \{day, poor light, pedestrians\}. Thus, lack of actual scene labels in the training samples significantly reduces classification performances for the simCLR based model. 

In Table \ref{res2} we observe that using the rbf-kernel based simCLR-based model and training on KITTI but testing on JAAD, all the frames get classified as the same major common class between both datasets \{day, poor light, city\}. Thus, the macro classification performances are low, however the weighted performances are higher owing to the class imbalance. Similarly for the same method training on JAAD while testing on KITTI, the test frames get categorized into the 2 overlapping categories representing lighting variations for city and pedestrians in daytime with shadows. Thus, the rule-based method however, shows consistent generalizability across varying training datasets.
\begin{table*}[ht!]
\caption{Performance of scene classification by varying training datasets.}
\begin{adjustbox}{width=6.4in,center}
\begin{tabular}{|c|c|c c|c|c c|c|}
\hline
Method&Accuracy&Macro Precision&Macro Recall&Macro f-score&Weighted Precision&Weighted Recall&Weighted f-score\\ \hline
Dataset&&Train: KITTI& Test: JAAD&&&&\\\hline
Rule-based&69.2&{\bf45.7}&{\bf48.7}&{\bf43.2}&{\bf92.6}&69.2&{\bf78.6}\\ 
simCLR-based&{\bf84.4}&16.9&20&18.3&71.2&{\bf84.4}&77.3\\ \hline
Dataset&&Train: JAAD& Test: KITTI&&&&\\ \hline
Rule-based&{\bf64.0}&{\bf58.6}&{\bf42.8}&{\bf40.7}&{\bf65.0}&{\bf64.0}&{\bf61.3}\\
simCLR-based&36.0&22.5&33.3&25.3&14.7&36.0&20.2\\ \hline
\end{tabular}
\label{res2}
\end{adjustbox}
\end{table*}

This analysis demonstrates that though simCLR is faster and more reliable when a critical mass of frames are annotated, it does not generalize well for new test datasets with new scene classes. Thus, for new use-cases and fields of view, the rule-based method is a good starting point followed by simCLR approach for speed up and accuracy.

\subsection{Filtering key frames}
Once each frame in video sequences is classified, the next task is to isolate frames with the highest uncertainty or previously unseen structural content that can then be annotated to re-train or fine-tune scene understanding algorithms. The fraction of all the test frames (in percentage) filtered by the two methods (training and testing on images from the same dataset) and the video sequences they belong to are shown in Table \ref{res3}. Additionally, we analyze the key-frames that are filtered by both methods to assess for structural agree-ability between the two methods. The frame filtering parameters for both methods are controlled to ensure similar fractions of filtered frames (less than 10\%). 
\begin{table}[ht!]
\caption{Key frame filtering performance from video sequences.}
\begin{adjustbox}{width=3.3in,center}
\begin{tabular}{|c|c|c|}
\hline
Dataset: Test JAAD&&\\ 
Method&\% Filtered frames&Video Sequences\\ \hline
Rule-based&301 (6.4)& 0024,0274,0275,0276,0277,0278,0279,0280,0281\\
simCLR-based&328 (7.0)&0024,0279,0280,0289,0304,0324\\
Rule-based $\cap$ simCLR&23 (0.5) &0024, 0279, 0280 \\ \hline
Dataset: Test KITTI&&\\ \hline
Rule-based&489 (7.6)&0002, 0003, 0007, 0008, 0009\\
simCLR-based&533 (8.35)&0003,0007, 0008, 0011, 0014, 0018, 0020\\ 
Rule-based $\cap$ simCLR&117 (1.7)& 0003, 0007, 0008\\\hline
\end{tabular}
\label{res3}
\end{adjustbox}
\end{table}

\begin{figure*}[ht]
    \centering
	\subfigure[Image with \{night, good light, city\} from sequence 0024, JAAD]
	{\includegraphics[width=0.45\textwidth, height=0.9in]{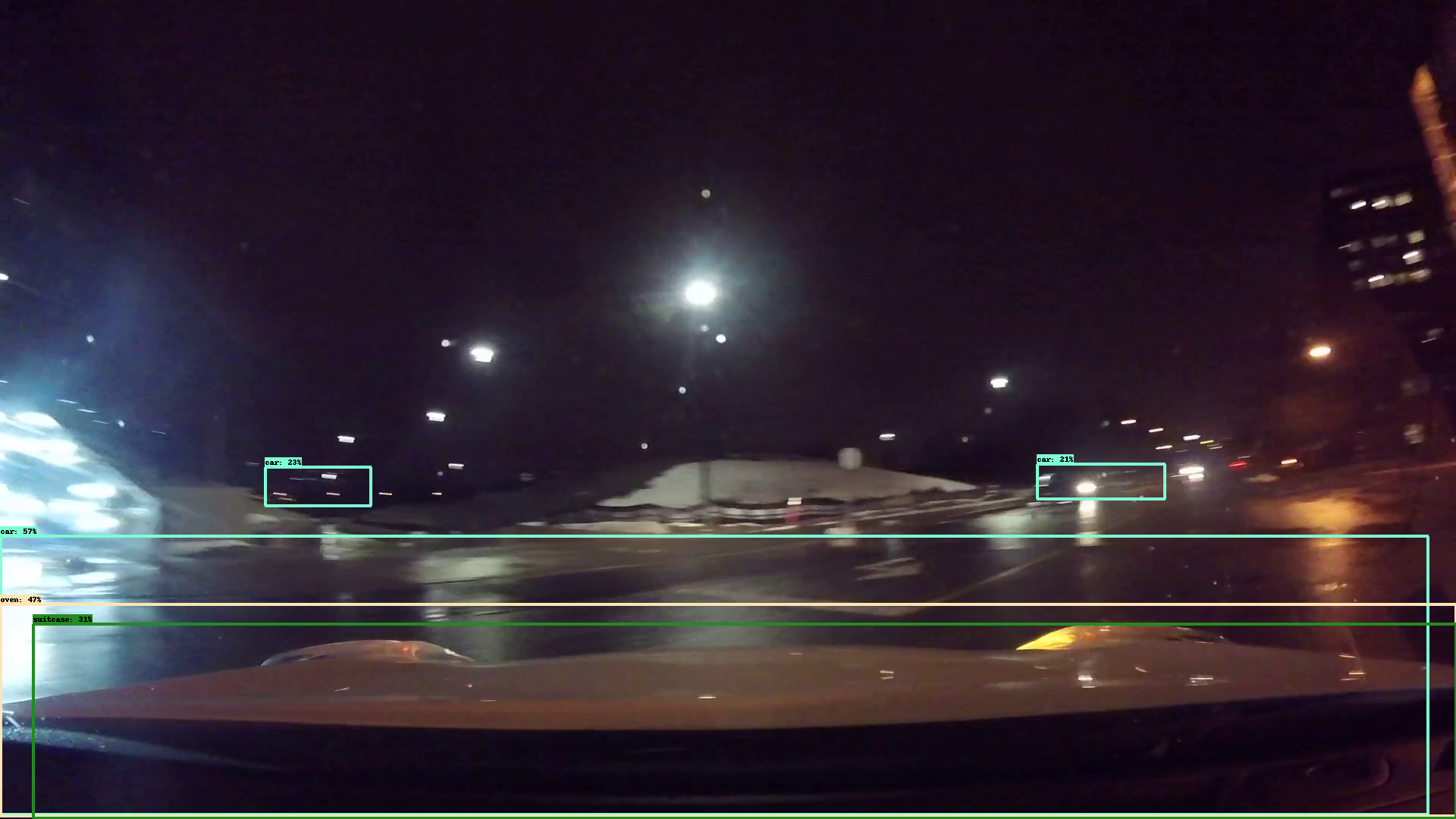}}
	\subfigure[Image with \{day, poor light, city\} from sequence 0273, JAAD]
	{\includegraphics[width=0.45\textwidth, height=0.9in]{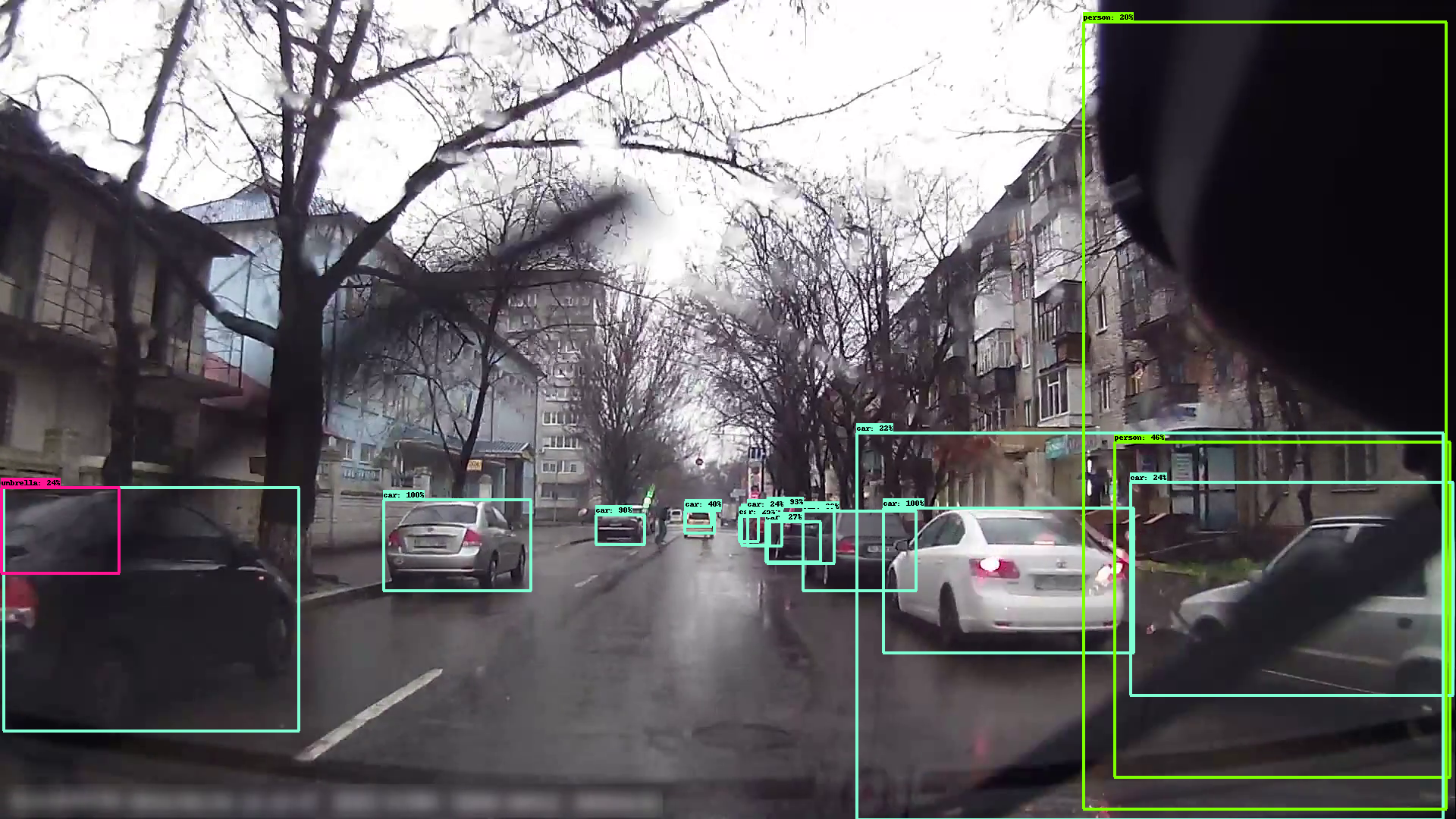}}
	\subfigure[Image with \{day, poor light, city\} from sequence 022, JAAD]
	{\includegraphics[width=0.45\textwidth, height=0.9in]{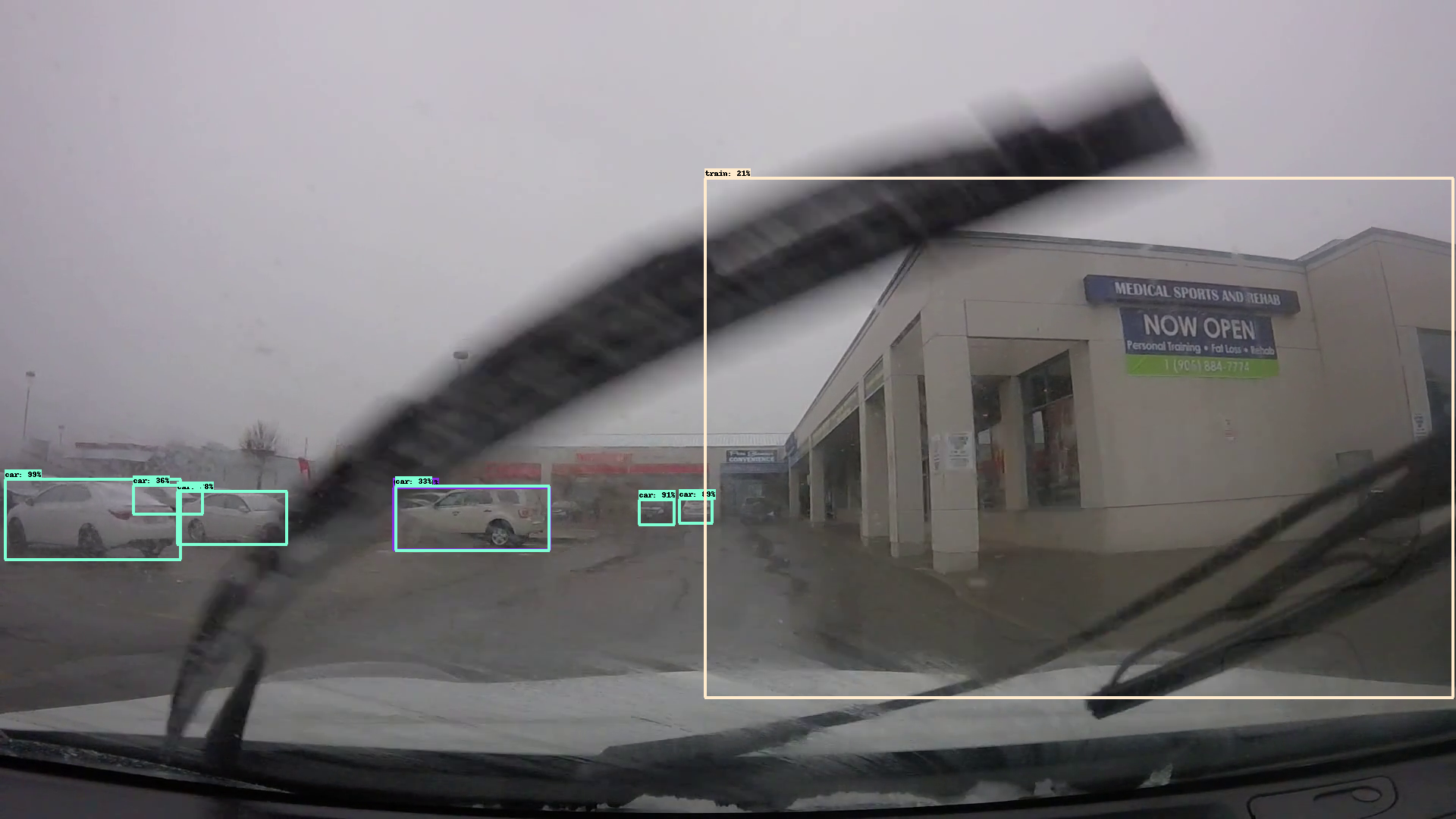}}
	\subfigure[Image with \{day, poor light, city\} from sequence 0276, JAAD]
	{\includegraphics[width=0.45\textwidth, height=0.9in]{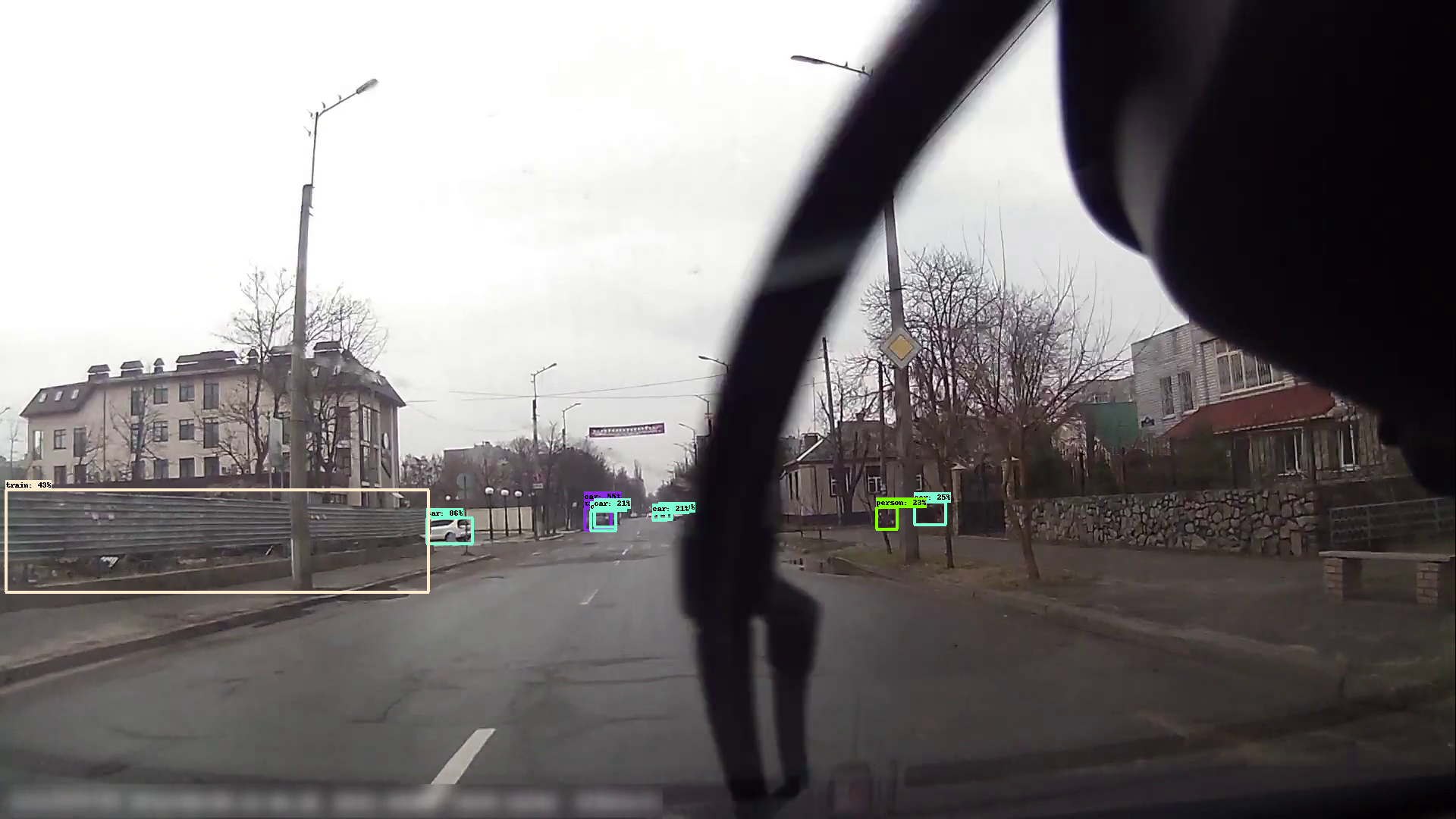}}
	\subfigure[Image with \{day, good light, city\}, sequence 0000, KITTI]
	{\includegraphics[width=0.45\textwidth, height=0.9in]{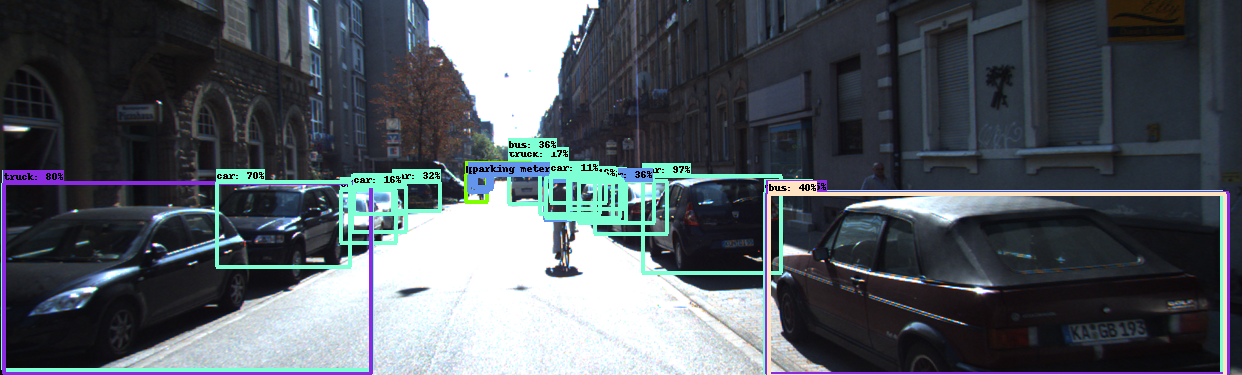}}
	\subfigure[Image with \{day, good light, parked-cars\}, sequence 0009, KITTI]
	{\includegraphics[width=0.45\textwidth, height=0.9in]{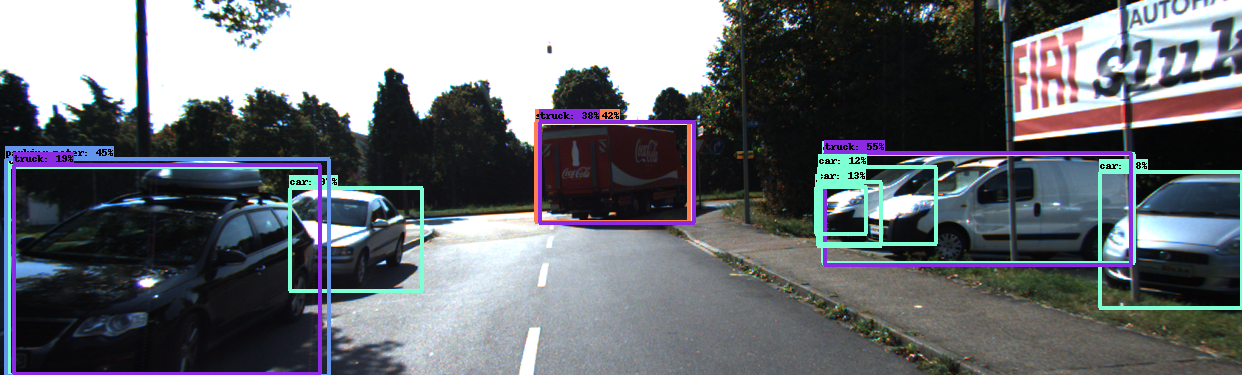}}
	\subfigure[Image with \{day, poor light, city\}, sequence 0019, KITTI]
	{\includegraphics[width=0.45\textwidth, height=0.9in]{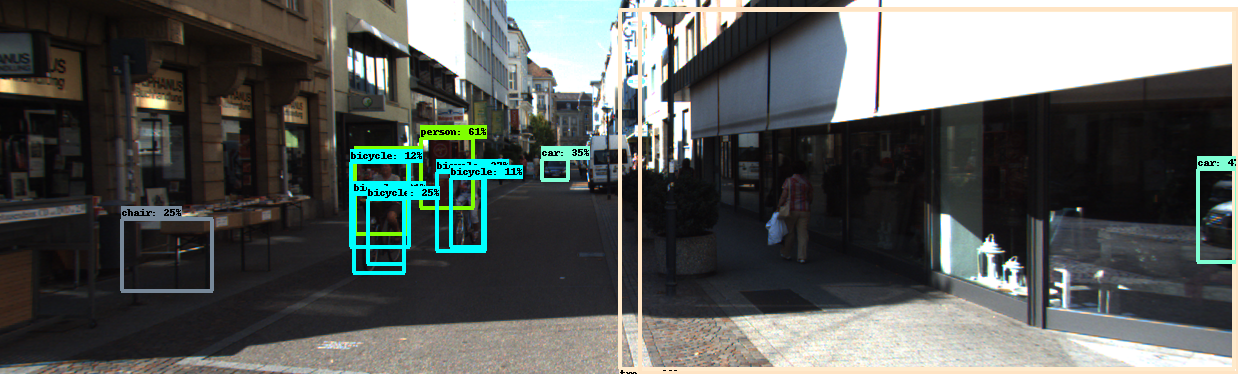}}
	\subfigure[Image with \{day, poor light, freeway\} from sequence 0020, KITTI]
	{\includegraphics[width=0.45\textwidth, height=0.9in]{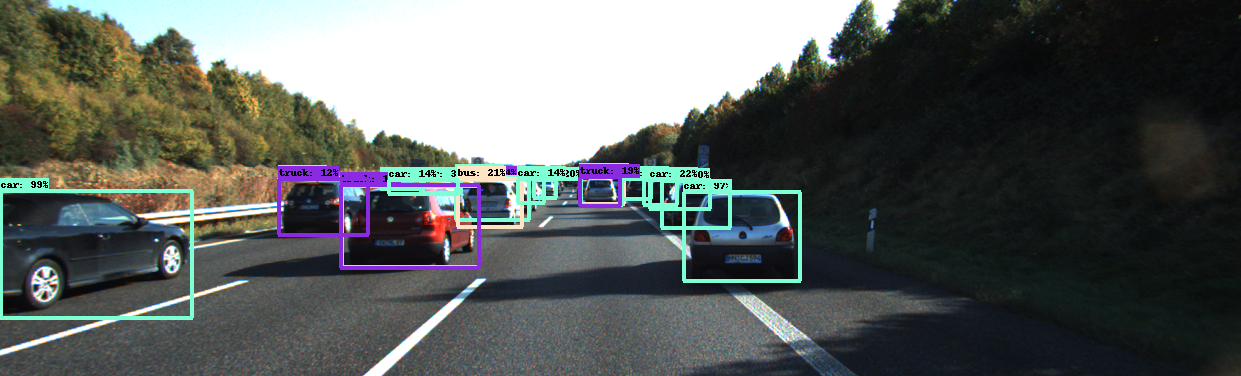}}
	\caption{Examples of images filtered by the rule-based baseline method.}\label{filtrule}
       \vspace{-0.3cm}
\end{figure*}

From Table \ref{res3} we observe that the rule-based method mostly isolates frames from video sequences with rain in the JAAD dataset. Running windshield wipers and droplets on the windscreen structurally alter the scene across frames. Some examples of key-frames filtered by this method are shown in Fig. \ref{filtrule}. Here, Fig. \ref{filtrule}(a) represents a blurry image that is eliminated by the frame quality detector/selection module. Also, the variations across boundary box probabilities in each of the poor lighting sequences are shown here.

\begin{figure*}[ht]
    \centering
	\subfigure[Night image (JAAD), where a pedestrian first appears (red box).]
	{\includegraphics[width=0.45\textwidth, height=0.9in]{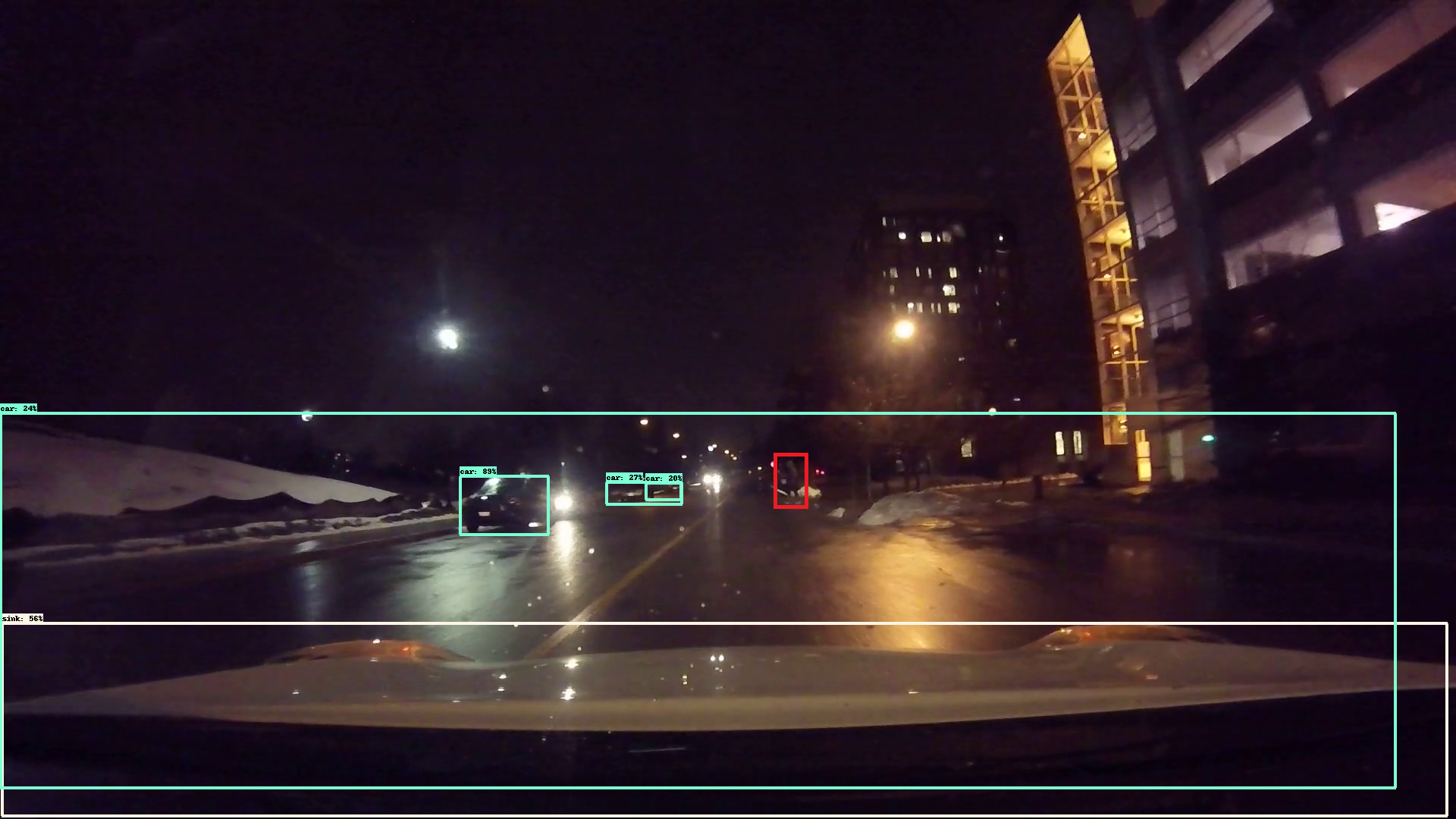}}
	\subfigure[Image from sequence 0279 in JAAD with several shadows.]
	{\includegraphics[width=0.45\textwidth, height=0.9in]{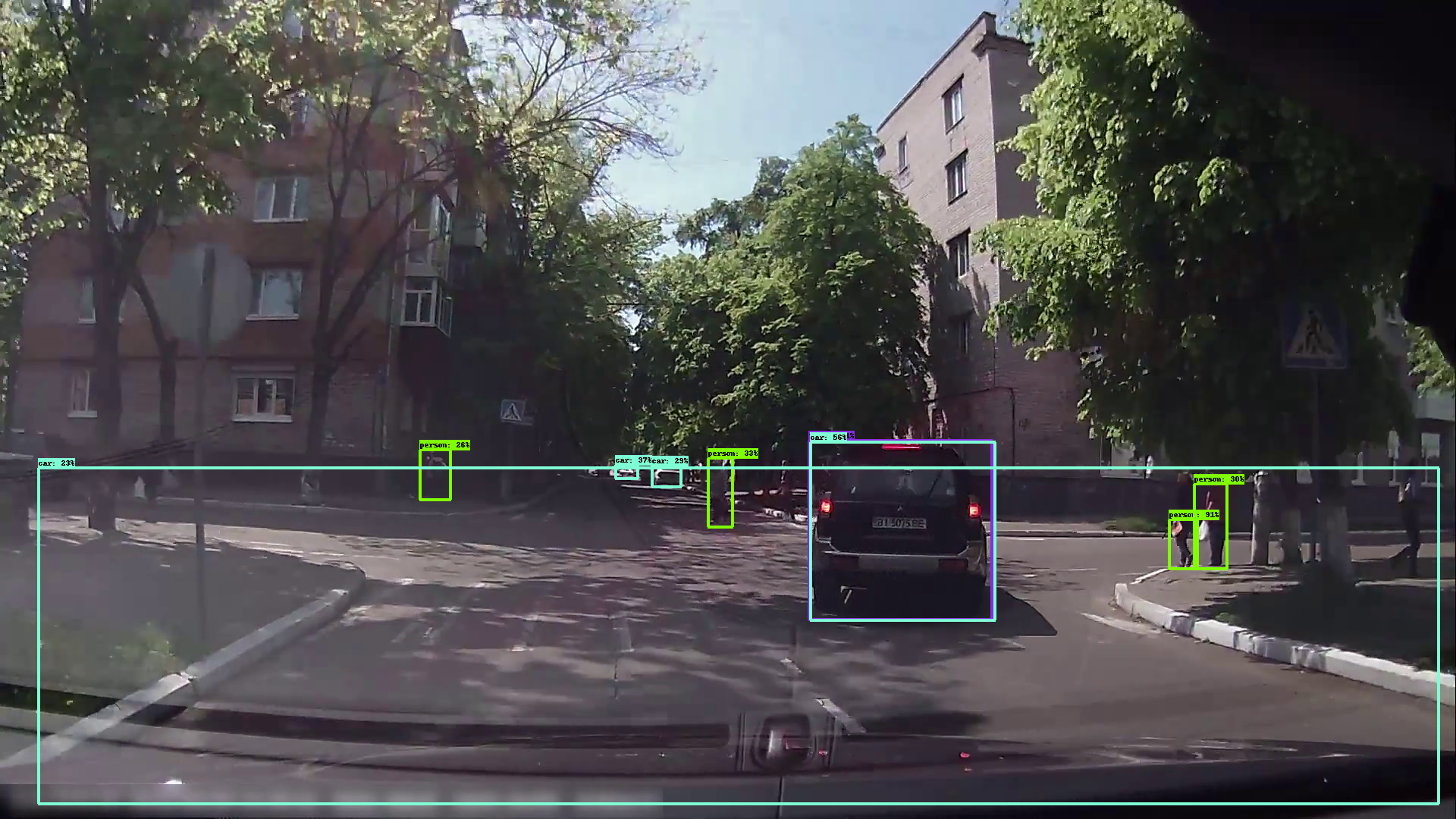}}
	\subfigure[Image from sequence 007 in KITTI with parked cars]
	{\includegraphics[width=0.45\textwidth, height=0.9in]{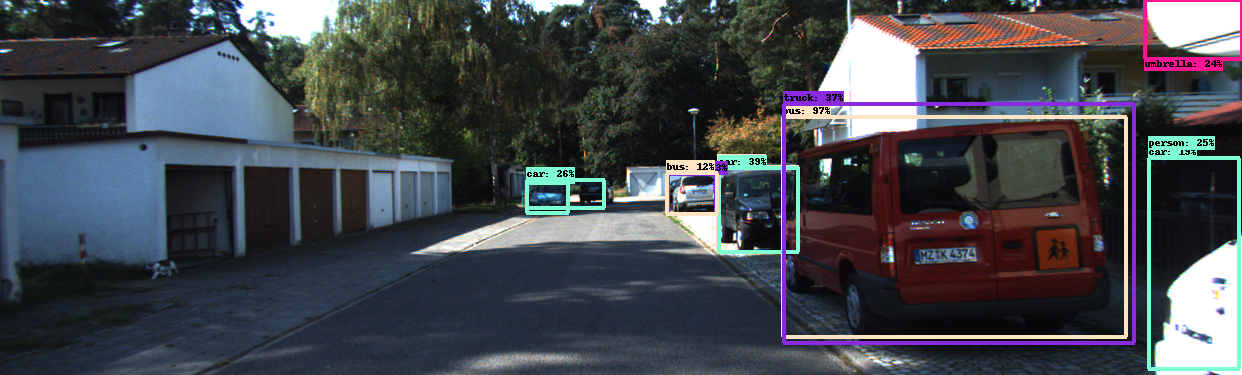}}
	\subfigure[Image from sequence 0011 in KITTI with poor light.]
	{\includegraphics[width=0.45\textwidth, height=0.9in]{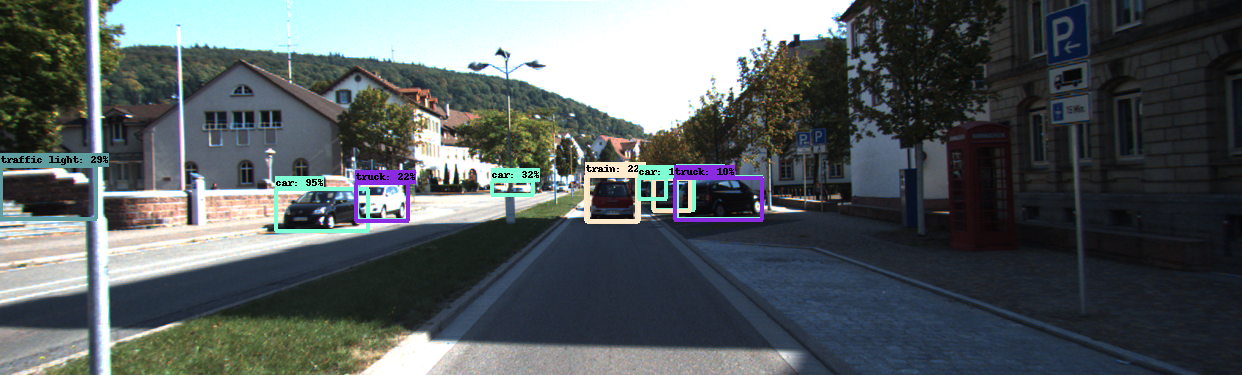}}
	\caption{Examples of key-frames filtered by the simCLR-based method.}\label{filtsim}
       \vspace{-0.3cm}
\end{figure*}

Additionally, from Table \ref{res3} we observe that the simCLR-based method filters frames from video sequences with shadows from trees and buildings in city regions in the JAAD dataset. Thus, all outdoor images with poor lighting combined with weather-related scenarios including snow and rain are found to contain structural uncertainty and to lie at cluster boundaries. For the KITTI dataset, most filtered key-frames correspond to scenes from parked cars in city residential areas, merging roads to freeways and freeway images with several cars and foliage. The most challenging decision boundary in the KITTI dataset therefore is to classify a freeway image with foliage from a city image with only parked cars in sight. Some examples of images filtered by this method are shown in Fig. \ref{filtsim}. Here, Fig \ref{filtsim}(a) represents a night-time frame where a pedestrian first appears. Filtering such frames and fine-tuning for such outlier situations will improve object detection tasks such as pedestrian detection at night. 
Finally, we observe that filtered frames common to both rule-based and simCLR methods for JAAD are the ones of night-time frames when pedestrians first appear ahead of the camera vehicle, from city scenes with shadows and poor lighting on a snowed road. Also, we observe a very small fraction of frames that are filtered by both methods. This occurs since the simCLR-based method filters frames that are structurally dissimilar to the training frames, such as snowed roads, rain, city to highway merging frames etc. On the other hand the rule-based method filters frames based on the uncertainty in content with respect to the previous and subsequent frames. Thus, based on the use-case and outcome requirements for key-frame filtering speed (using simCLR-based method) vs. generalizability across structural content (using the rule-based method), using either method for isolating a minimal training data batch is effective for scene understanding tasks.

To further assess the importance of filtered key-frames for transfer learning, we further fine-tune the Faster-RCNN model using the filtered frames obtained from the simCLR model. The object detection performances on the JAAD test sequences (pedestrian only) are: mean average precision (mAP) of 70.91 and mean intersection over union (mIOU) of 15.4. On the KITTI data set (vehicles and pedestrians only), the fine-tuned object detector has mAP of 81.7 and mIOU of 17.8, which is comparable to state-of-the-art models that use large volumes of training images for object detection \cite{KITTItracking}.

\section{Conclusions and Discussion}
Video data processing and tasks often require sifting through several hours worth of video to create an efficient data pipeline that will then be used by modeling and deployment pipelines in an operationalized machine learning system. These video data pipelines rely on efficient tagging, storage and retrieval of video frames to extract relevant data batches for the modeling pipeline. For instance, fine-tuning an autonomous drive perception system for highway perception or night-time object detection requires training on relevant image samples that can be facilitated only if raw input frames were tagged during storage for scene class, lighting conditions or foreground content etc. Additionally, isolating relevant \textit{key-frames} that improve scene understanding algorithms and sending those images for annotation followed by model fine-tuning in batches is key to continuous data-centric machine learning deployment. In this work we present two methods to automate the process of tagging each frame in video sequences and filtering key-frames based on structural content to capture variability in image content from a minimal training dataset.

The first structural content based method requires minimal parameterization and relies on foreground objects and structural uncertainty information across frames for scene classification with 64-72\% accuracy across varying training datasets. This method is most generalizable but suffers from a high processing time of 0.32-0.5 seconds per image. The second method relies on high dimensional encoding using a self-supervised deep learning model followed by label spreading that achieves 66-93\% accuracy of frame classification by training on similar images only. This method is extremely fast with processing times of 0.035 seconds per frame. 

Future works will be directed towards combining the deep-learning encoding with the sequential information from neighboring frames to further enhance the automated frame tagging methods. Additionally, the proposed methods can be extended for visual inspection and indoor camera systems for outlier and threat detection tasks in future.

\bibliographystyle{IEEEtran}
\bibliography{papers}

\end{document}